\pdfoutput=1 % tell arXiv to build with pdfLaTeX (figures are PNG)
\documentclass{article}

\PassOptionsToPackage{numbers}{natbib}

\usepackage[preprint]{neurips_2026}

\usepackage[utf8]{inputenc} % allow utf-8 input
\usepackage[T1]{fontenc}    % use 8-bit T1 fonts
\usepackage{hyperref}       % hyperlinks
\usepackage{url}            % simple URL typesetting
\usepackage{booktabs}       % professional-quality tables
\usepackage{amsfonts}       % blackboard math symbols
\usepackage{amsmath}        % math environments
\usepackage{nicefrac}       % compact symbols for 1/2, etc.
\usepackage{microtype}      % microtypography
\usepackage{xcolor}         % colors
\usepackage{graphicx}       % figures
\hypersetup{hidelinks}      % keep cross-references unobtrusive in print

\title{When Do Attention-Head Ablations Support Causal Claims? \\
Projection-Level Confounds, Floor Effects, and Matched Controls}

\author{%
  Juli Huang \\
  Stanford University \\
  \texttt{julih@stanford.edu}
}

\begin{document}

\maketitle

\begin{abstract}
Attention-head ablation, zeroing a head and measuring the resulting
change in task performance, is a standard way to infer which
components of a language model are causally responsible for a behavior.
We show, using GPT-2 small as a case study, that this recipe is fragile
in ways that are rarely checked. First, we find that a natural
implementation of ``zero out head $h$'' (overwriting a channel slice of
the attention block's \emph{returned} output) does not ablate a single
head at all: the block's output projection mixes contributions across
heads into output channels before the block returns, so this
intervention zeroes an arbitrary slice of the already-mixed
residual-stream update. On a controlled induction-style copying task, this
naive method is essentially uncorrelated with a corrected,
pre-projection ablation (Pearson $r=0.057$) and selects a completely
disjoint top-5 set of ``important'' heads (0/5 overlap). Second, using
the corrected method, we show that accuracy cannot detect further
degradation at a behavioral floor (0.0\% on an off-target arithmetic
task). Near a ceiling (99.5\% on the controlled induction-style copying
task), it can detect large degradations but may miss confidence changes
that do not cross the decision boundary. Log-probability of the gold
continuation stays graded in both regimes. Third, with a discovery/held-out split and 1{,}000 matched
random-head and layer-matched-head control draws, the per-head effect
ranking is stable across the two splits (Spearman $\rho=0.974$ over all
144 heads), and the top-5 discovery-selected heads' mean held-out effect
substantially exceeds both control distributions (Monte Carlo
$p=0.001$). Evidence for task specificity is
not robust on GPT-2: the primary run is suggestive ($p=0.078$), while a
single pre-committed larger rerun gives $p=0.274$. On DistilGPT2, the
intervention-semantic and matched-control findings replicate, and the
arithmetic floor recurs. Copying accuracy is lower (90.0\%), so GPT-2's
near-ceiling example does not replicate. A single-head ablation result
is not self-certifying: it requires verifying the intervention's
projection semantics, a continuous metric away from floor/ceiling, and
matched controls evaluated on held-out data.
\end{abstract}

\section{Introduction}
\label{sec:intro}

Interpretability researchers frequently ablate an individual attention
head, zeroing its contribution and re-measuring task performance, and
interpret a resulting drop as evidence that the head is causally
responsible for some computation. Whether this inference is valid
depends on choices that are rarely audited: what, mechanically, gets
zeroed relative to the attention block's linear mixing structure;
whether the evaluation metric has room to move at all; and whether the
identified heads are validated against an appropriate null rather than
just reported as the top of a ranking. This paper asks a narrow,
mechanical question: \textbf{under what conditions does single-head
zero-ablation support a defensible causal interpretation}, in the sense
of an effect that exceeds matched controls on held-out data?

We answer it by auditing and fixing a head-ablation implementation,
designing a controlled induction-style copying task with behavioral
headroom alongside an off-target floor task, and running a discovery/held-out
pipeline with random and layer-matched controls, on GPT-2 small
(124M parameters, 12 layers $\times$ 12 heads). We report three
findings, one per research question (RQ0--RQ2), matching the
Section~\ref{sec:results} subsections below.

\textbf{(1, RQ0) The ablation implementation matters.} A hook registered on
the whole attention module's output, a natural way to write ``zero out
head $h$'', zeroes channels \emph{after} the block's output projection
has already mixed all heads together, so it does not isolate one head's
contribution. On our controlled copying task, this naive method's per-head effect
vector is essentially uncorrelated with a corrected, pre-projection
version (Pearson $r=0.057$), and the two methods' top-5 ``important''
heads share zero heads in common.

\textbf{(2, RQ1) Binary accuracy loses information at floor and near ceiling.} GPT-2
small scores 0.0\% on an off-target arithmetic task and 99.5\% on our
on-target copying task. At the floor, accuracy cannot register further
degradation. Near the ceiling, it can register a sufficiently large
drop but may hide confidence changes that leave the predicted token
unchanged. Teacher-forced log-probability of the gold continuation
remains graded in both regimes.

\textbf{(3, RQ2) With proper controls, five heads show a held-out-replicated
intervention effect that exceeds random and layer-matched baselines, but their
task-specificity is not robust.} Using log-probability, a discovery/held-out
split, and 1{,}000-draw matched control distributions, the per-head
effect ranking is stable across splits (Spearman $\rho=0.974$ over all
144 heads), and the top-5 discovery-selected heads' mean held-out effect
clears both control distributions decisively (Monte Carlo $p=0.001$
each). Their advantage on the target task over an off-target
arithmetic control is only weakly distinguishable from a random-head
baseline in the primary run (Monte Carlo $p=0.078$) and is weaker in a
single pre-committed $N=500$ rerun ($p=0.274$).

Our contribution is a worked, fully reproducible example (code, seeds,
and every number traceable to a saved result file) showing that a
positive \emph{or} null single-head ablation result is not
self-certifying: a causal interpretation additionally requires (i)
verifying the intervention's projection semantics, (ii) a continuous
metric that is not saturated, and (iii) matched controls evaluated on
data not used for head selection.

\section{Why naive head ablation can mislead}
\label{sec:background}

GPT-2's attention block computes a per-head output for each head,
concatenates the per-head outputs, and applies a single dense output
projection ($W_O$, \texttt{c\_proj} in the standard \texttt{transformers}
implementation) before adding the result to the residual stream.
\citet{elhage2021framework} describe $W_O$ as half of each
head's output-value (OV) circuit. Because $W_O$ spans the concatenated head dimension,
each output channel can receive contributions from all heads.
A hook that zeroes a contiguous, head-sized slice of the block's
\emph{returned} output (registered as a forward hook on the whole
attention module) therefore does not remove one head's contribution. It
removes an arbitrary linear combination of all heads' contributions to a
subset of output channels, after mixing has already happened. The
correct intervention instead zeroes the corresponding slice of $W_O$'s
\emph{input}, i.e., the concatenated, not-yet-mixed per-head values,
which requires a pre-hook on the projection itself rather than a
post-hook on the block.

We did not construct this as a strawman: it is exactly the
implementation we found while auditing a pre-existing, previously
unaudited attention-head-ablation implementation, and we use it as
the ``naive'' method in our RQ0 comparison (Section~\ref{sec:results}).
The methods use the same nominal head index, model, and input. Their
activation interventions differ only in where they are applied within
the block.

Separately, prior work has repeatedly cautioned that single-component
interventions and attention weights are easy to over-interpret without
controls: \citet{jain2019attention} show attention
weights alone are often uncorrelated with feature importance, and
\citet{michel2019sixteen} show many heads can be pruned
post hoc without hurting overall performance, i.e., that a head being
prunable does not mean it was doing nothing task-specific. More recent
circuit-style work explicitly validates causal claims about specific
components against matched baselines and held-out data
\citep{meng2022locating,wang2023interpretability,conmy2023towards}. Our contribution is
narrower and more mechanical: we isolate the projection-level
implementation confound specifically, and combine it with a
floor/ceiling-aware metric and a minimal control battery in one
end-to-end worked example.

\section{Experimental design}
\label{sec:design}

\textbf{Model.} GPT-2 small (124M parameters, 12 layers, 12 heads/layer,
144 heads total), the standard Hugging Face \texttt{transformers}
implementation (v4.57.3, torch~2.8.0), run on CPU.

\textbf{Tasks.} \emph{Controlled induction-style copying} (target task, expected
headroom): synthetic four-word sequences of the form ``$w_1\ w_2\ w_3\
w_4$ . $w_1\ w_2\ w_3$'' $\to$ ``\ $w_4$'', with $w_1,\dots,w_4$ drawn
without replacement from a pool of common English words verified to
tokenize as exactly one byte-pair-encoding (BPE) token, a minimal instance of the
induction-head phenomenon described by
\citet{olsson2022induction} and traced to two-layer head
composition by \citet{elhage2021framework}. We generated
$N=200$ examples (seed 0) and split them 50/50 into a \emph{discovery}
set (used only to select candidate heads) and a \emph{held-out} set
(used only to evaluate them), fixed before any head effect was computed.
\emph{Arithmetic} (off-target control task): ``Q: $a$+$b$='' $\to$ ``\
$a{+}b$'', $a,b$ drawn uniformly from $[1,20]$, $N=200$ (seed 1), used
only to test task-specificity of already-selected heads, never for
selection. Every generated example is checked so that the question's
tokenization is an exact token-level prefix of question+answer (BPE
merges are not always prefix-stable across concatenation). Examples
failing this check are dropped and regenerated.

\textbf{Metrics.} \emph{Teacher-forced exact match}: the greedy argmax
token at every answer position must equal the gold token (binary).
\emph{Teacher-forced log-probability}: $\sum_t \log p(\text{gold
token}_t \mid \text{prefix})$ over the answer's tokens (continuous,
natural log). Both require a single forward pass per batch (no
autoregressive generation).

\textbf{Ablation methods.} \texttt{naive\_post\_projection}: a forward
hook on the attention module's output, zeroing a head-sized channel
slice of the returned (already-projected) tensor. This is invalid and
used only for the RQ0 comparison. \texttt{head\_ablation}: a forward
pre-hook on the output projection's input, zeroing the corresponding
pre-mix slice. This is the only method used for any causal claim in
Sections~\ref{sec:rq1}--\ref{sec:rq2}. These semantics are asserted by
executable tests that run before the pipeline and ship with the code: on a
small GPT-2 configuration they check that the pre-hook zeroes exactly the
requested head's slice of the projection input and leaves every other
channel unchanged, that removing the hook exactly restores the unablated
output, and that the naive hook alters the logits while leaving that
projection input untouched.

\textbf{Head selection.} For all 144 heads, we compute the mean
$\Delta = \overline{\text{log-prob}}_{\text{baseline}} -
\overline{\text{log-prob}}_{\text{ablated}}$ over a data split, using
one batched forward pass per head (not per example). Candidate heads
are the top-5 by $\Delta$ on the discovery split only.

\textbf{Controls.} \emph{Random-head control}: 1{,}000 draws of 5 heads
sampled uniformly without replacement from all 144. \emph{Layer-matched
control}: 1{,}000 draws that preserve the candidate set's size and layer
composition while replacing every candidate identity, drawing each
layer's replacements without replacement from that layer's
non-candidate heads. Both exclusions matter whenever one layer supplies
several candidates, as ours does: replacing each candidate independently
would let a single head stand in for several candidates, and would let a
candidate head enter its own control distribution. Both controls are
built by resampling the already-computed 144-head $\Delta$ vector, so
they cost no additional forward passes.

\textbf{Statistics.} Where applicable, example-level point estimates are
reported with percentile-bootstrap 95\% confidence intervals (CIs;
2{,}000 resamples over examples). Control ranges are the 5th--95th
percentiles of 1{,}000 draws. One-sided Monte Carlo $p$-values use the
finite-sample correction $(k+1)/(B+1)$, where $k$ of $B=1{,}000$ control
draws match or exceed the candidate statistic. Rank stability across
splits is measured with Spearman correlation over all 144 heads.

\textbf{Reproducibility.} Seeds (induction generation~0, arithmetic
generation~1, control sampling~42), package versions, and the experiment
configuration are recorded with every saved result. The full pipeline
(\texttt{scripts/run\_neurips\_experiments.py}) reruns end-to-end in
198.99~s for the primary GPT-2 configuration on one 10-core Apple M5 CPU
with 24~GB unified memory and no GPU (peak resident memory 1.94~GiB),
57.62~s for DistilGPT2, and 428.56~s for GPT-2 at $N=500$ (3.22~GiB).
Each of the three configurations was then rerun independently; all three
reproduced their saved result files exactly, byte for byte, apart from
the recorded timestamp. The software environment, model revisions,
file hashes, storage, and verification commands are recorded in
\texttt{neurips/experiment\_results/reproducibility\_provenance.md}. Every
number in this paper is read from a result JSON file.

\section{Results}
\label{sec:results}

\subsection{RQ0: Does the ablation implementation matter?}
\label{sec:rq0}

On the controlled copying-task discovery split, the naive and correct per-head
$\Delta$ vectors are nearly uncorrelated (Pearson $r=0.057$, Spearman
$\rho=0.110$; scatter plot in Appendix~\ref{app:rq0}), with a mean
absolute per-head disagreement of $0.067$ nats, comparable in
magnitude to the candidate heads' own held-out effect ($0.412$ nats,
Section~\ref{sec:rq2}). The top-5 heads selected by each method share
\textbf{zero} heads. The methods use the same nominal head index,
model, and input, but apply the activation intervention at different
points in the block (Section~\ref{sec:background}).
This is not a hypothetical failure mode: the naive intervention is a
natural, easy-to-write hook, and we drew it directly from a
pre-existing, previously unaudited implementation rather than
constructing it to fail.

\subsection{RQ1: Metric behavior at floor and near ceiling}
\label{sec:rq1}

GPT-2 small is at a behavioral floor on the arithmetic control task
(0.0\% teacher-forced exact match, 95\% CI $[0.0\%, 0.0\%]$, $N=200$)
and near a behavioral ceiling on the controlled induction-style
copying task (99.5\%, CI $[98.5\%,100.0\%]$). At the floor, binary
accuracy cannot detect further degradation. Near the ceiling, it can
detect a sufficiently large damaging effect, but a substantial
confidence decrease may leave every argmax prediction unchanged. Log-probability
remains graded: mean log-prob is $-7.44$ nats (CI
$[-7.53,-7.35]$) on arithmetic and $-0.83$ nats (CI $[-0.90,-0.76]$) on
the copying task (Table~\ref{tab:main}, Figure~\ref{fig:main}A), and, as shown
next, individual-head
ablation moves it substantially and reproducibly. This is why we select
and validate candidate heads using log-probability rather than accuracy
throughout.

\subsection{RQ2: Specificity of selected-head intervention effects}
\label{sec:rq2}

\begin{figure}[t]
  \centering
  \includegraphics[width=\linewidth]{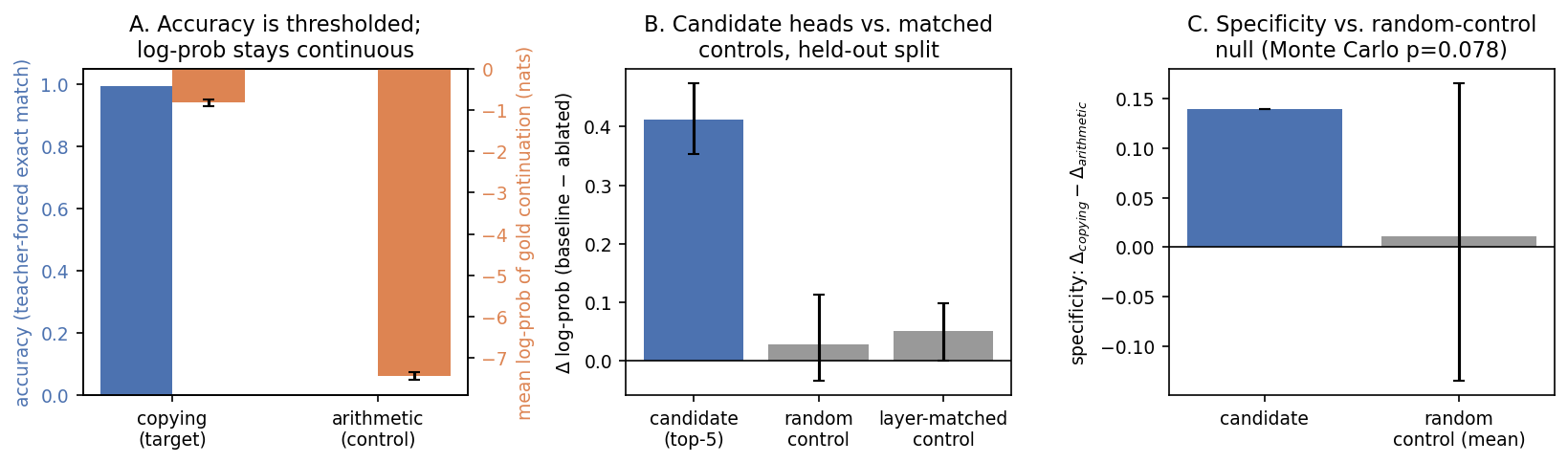}
  \caption{\textbf{A.} Accuracy cannot show further degradation at the
  arithmetic floor and may hide confidence changes near the copying-task
  ceiling. Log-probability stays continuous.
  \textbf{B.} The top-5 candidate heads' held-out effect far exceeds
  matched random-head and layer-matched control distributions (error
  bars: 95\% CI for the candidate, 5th--95th percentile for controls).
  \textbf{C.} Specificity (target-task minus control-task effect) is
  larger for candidates than the random-control mean, but the
  difference is only suggestive (Monte Carlo $p=0.078$).}
  \label{fig:main}
\end{figure}

The top-5 discovery-selected heads are Layer~0 Head~0, Layer~0 Head~7,
Layer~1 Head~3, Layer~11 Head~0, and Layer~0 Head~10 (four in the first
two layers, one in the last layer). Two distinct pieces of evidence
support them. First, the per-head effect ranking is highly stable across
the discovery and held-out splits (Spearman $\rho=0.974$ over all 144
heads), so head ordering is not an artifact of the discovery sample.
Second, the candidates' mean held-out effect, $0.412$ nats (CI
$[0.353,0.474]$), is far outside both the matched random-head control
distribution (mean $0.029$, 5th--95th percentile range
$[-0.033,0.113]$; Monte Carlo $p=0.001$,
0/1{,}000 draws matched or exceeded the candidates) and the
layer-matched control distribution (mean $0.051$, 5th--95th percentile
range $[0.001,0.098]$;
Monte Carlo $p=0.001$; Figure~\ref{fig:main}B). This establishes a held-out-replicated
intervention effect on the controlled copying task that exceeds what an arbitrary
head, or an arbitrary head from the same layers, would produce. Whether
that effect is specific to the copying task requires the separate contrast below.

Task-specificity is weaker (Figure~\ref{fig:main}C). The specificity
statistic (held-out copying-task $\Delta$ minus arithmetic-control
$\Delta$) is $0.139$ nats
for the candidates, versus a random-control distribution with mean
$0.011$ and 5th--95th percentile range $[-0.130,0.163]$. The Monte Carlo
$p$-value is $0.078$, a
positive trend that does not clear a conventional 0.05 threshold. We
report this as suggestive rather than conclusive: the candidates' held-out
advantage is larger on the target task than on the off-target control
task, but we cannot rule out that part of it reflects generic disruption
rather than a computation specific to the copying task.

\begin{table}[t]
  \caption{Summary of all RQ1/RQ2 quantities. All values are computed
  from \texttt{gpt2\_results.json} (Section~\ref{sec:design}). CIs are
  95\% percentile bootstrap over examples except where noted.
  ``Control range'' is the 5th--95th percentile of 1{,}000 control
  draws, not a CI.}
  \label{tab:main}
  \centering
  \small
  \begin{tabular}{lrrl}
    \toprule
    Quantity & Value & Interval & $N$ / draws \\
    \midrule
    Copying-task accuracy (ceiling) & 0.995 & [0.985, 1.000] & 200 \\
    Arithmetic accuracy (floor) & 0.000 & [0.000, 0.000] & 200 \\
    Copying-task log-prob (nats) & $-0.83$ & [$-0.90$, $-0.76$] & 200 \\
    Arithmetic log-prob (nats) & $-7.44$ & [$-7.53$, $-7.35$] & 200 \\
    Candidate $\Delta$, held-out & 0.412 & [0.353, 0.474] & 100 \\
    Random-control $\Delta$, held-out & 0.029 & [$-0.033$, 0.113]$^\dagger$ & 1000 \\
    Layer-matched $\Delta$, held-out & 0.051 & [0.001, 0.098]$^\dagger$ & 1000 \\
    Rank stability (disc.\ vs.\ held-out, $\rho$) & 0.974 & -- & 144 heads \\
    Candidate specificity & 0.139 & -- & -- \\
    Random-control specificity & 0.011 & [$-0.130$, 0.163]$^\dagger$ & 1000 \\
    \bottomrule
  \end{tabular}
  \vspace{0.3em}
  \\
  \raggedright \footnotesize Monte Carlo $p$(candidate $\geq$ random control, held-out $\Delta$) $=0.001$; \quad
  $p$(candidate $\geq$ layer-matched, held-out $\Delta$) $=0.001$; \quad
  $p$(candidate $\geq$ random control, specificity) $=0.078$. $^\dagger$5th--95th percentile.
\end{table}

\section{Discussion and limitations}
\label{sec:discussion}

\textbf{Scope of the causal claim.} We do not claim these five heads
``compute induction'' in a mechanistic sense. Using the causal-claim
ladder (correlation $<$ decodability $<$ intervention effect $<$
specific causal contribution $<$ mechanistic role), our evidence
supports a held-out-replicated intervention effect that exceeds matched
controls (the third rung). The inconsistent GPT-2 specificity contrast
does not support the fourth rung for that model.
This is not a mechanistic account of what the heads compute. That would
require additional evidence such as activation patching, input-dependence
analysis, or directional interventions.

\textbf{Zero-ablation is off-distribution.} Setting a head's contribution
to exactly zero produces an activation the model never saw during
training. An observed effect could partly reflect this distribution
shift rather than the specific removal of a computation. Mean-ablation
or resampling-based interventions would be a natural robustness check
that we did not run.

\textbf{Model scope.} On DistilGPT2, the naive and correct effects are
again near-uncorrelated ($r=0.001$, 0/5 top-head overlap), the arithmetic
floor recurs, and the selected heads exceed both matched controls and the
specificity control ($p=0.001$ each; Appendix~\ref{app:distilgpt2}). Copying
accuracy is lower (90.0\%), so the near-ceiling example is specific to
GPT-2. Both models share one architecture family, and we did not test an
unrelated architecture.

\textbf{Task specificity is not robust on GPT-2.} The primary run gives
$p=0.078$. A pre-committed $N=500$ rerun preserves the held-out advantage
over both matched controls ($p=0.001$ each), but specificity weakens to
$0.056$ nats ($p=0.274$). This argues against stable GPT-2 task specificity.
We did not try further control tasks or thresholds.

\textbf{Single discovery/held-out split.} Candidate heads were selected
from one fixed 50/50 partition of the copying-task data. We did not test
sensitivity of the candidate set to alternative random partitions.

\textbf{Broader implication.} Each check changed the conclusion: the naive
hook selected disjoint heads, accuracy concealed different information at
the floor and near ceiling, and the held-out specificity contrast blocked a
task-specificity claim that the larger GPT-2 run does not support.

\section{Conclusion}
\label{sec:conclusion}

A head ablation is not self-certifying evidence of a specific causal role.
Hook placement can ablate the wrong quantity, accuracy is censored at a floor
and coarse near a ceiling, and held-out controls separate a replicating effect
from arbitrary disruption. Task specificity can still fail to replicate.

\bibliographystyle{unsrtnat}
{\small
\bibliography{references}
}

\newpage
\appendix

\section{Supplementary figure: RQ0 per-head comparison}
\label{app:rq0}

\begin{figure}[h]
  \centering
  \includegraphics[width=0.6\linewidth]{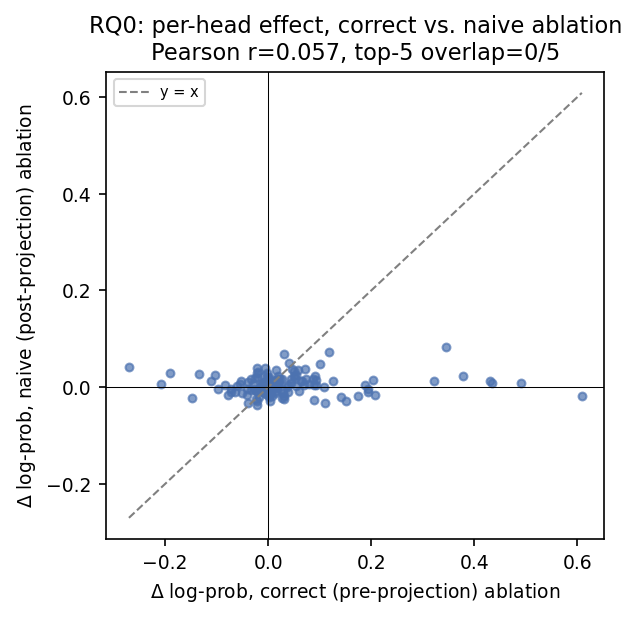}
  \caption{Per-head $\Delta$ log-prob under the correct (pre-projection)
  vs.\ naive (post-projection) ablation method, all 144 heads,
  controlled copying-task discovery split. A valid replication of the naive method
  would lie near the $y=x$ line. Instead the two methods are nearly
  uncorrelated (Pearson $r=0.057$).}
  \label{fig:rq0}
\end{figure}

Figure~\ref{fig:rq0} plots the per-head effect (correct pre-projection
ablation vs.\ naive post-projection ablation) for all 144 heads on the
controlled copying-task discovery split, corresponding to the summary statistics in
Section~\ref{sec:rq0}.

\section{Replication on a second model (DistilGPT2)}
\label{app:distilgpt2}

As a robustness check on the ``single model family'' limitation
(Section~\ref{sec:discussion}), we reran the identical pipeline
(Section~\ref{sec:design}), unchanged except for the model, on
DistilGPT2 (6 layers, 12 heads/layer, 72 heads total). All data,
metrics, ablation methods, splits, and control procedures are exactly
as in Section~\ref{sec:design}, and only the model differs. Results are in
Table~\ref{tab:distilgpt2}.

\begin{table}[h]
  \caption{DistilGPT2 replication of Table~\ref{tab:main}, with the same
  procedure and $N$s. The intervention-semantic and matched-control
  findings replicate, and the arithmetic floor recurs. Copying accuracy
  is lower (90.0\%), so GPT-2's near-ceiling example does not replicate.
  The specificity comparison gives Monte Carlo $p=0.001$.}
  \label{tab:distilgpt2}
  \centering
  \small
  \begin{tabular}{lrrl}
    \toprule
    Quantity & Value & Interval & $N$ / draws \\
    \midrule
    Copying-task accuracy & 0.900 & [0.855, 0.940] & 200 \\
    Arithmetic accuracy (floor) & 0.000 & [0.000, 0.000] & 200 \\
    Copying-task log-prob (nats) & $-0.96$ & [$-1.07$, $-0.85$] & 200 \\
    Arithmetic log-prob (nats) & $-7.38$ & [$-7.50$, $-7.25$] & 200 \\
    RQ0: Pearson $r$(naive, correct) & 0.001 & -- & 72 heads \\
    RQ0: top-5 head overlap & 0/5 & -- & -- \\
    Candidate $\Delta$, held-out & 0.586 & [0.511, 0.661] & 100 \\
    Random-control $\Delta$, held-out & 0.080 & [$-0.026$, 0.231]$^\dagger$ & 1000 \\
    Layer-matched $\Delta$, held-out & 0.054 & [$-0.017$, 0.128]$^\dagger$ & 1000 \\
    Rank stability ($\rho$, disc.\ vs.\ held-out) & 0.957 & -- & 72 heads \\
    Candidate specificity & 0.492 & -- & -- \\
    Random-control specificity & 0.023 & [$-0.159$, 0.222]$^\dagger$ & 1000 \\
    \bottomrule
  \end{tabular}
  \vspace{0.3em}
  \\
  \raggedright \footnotesize Monte Carlo $p$(candidate $\geq$ random control, held-out $\Delta$) $=0.001$; \quad
  $p$(candidate $\geq$ layer-matched, held-out $\Delta$) $=0.001$; \quad
  $p$(candidate $\geq$ random control, specificity) $=0.001$. $^\dagger$5th--95th percentile.
\end{table}

The candidate heads found on DistilGPT2 (Layer~0 Heads~1, 7, 9;
Layer~3 Head~2; Layer~5 Head~0) are, as expected, different indices
from GPT-2 small's. The two models are different parameter sets, so
this is not evidence the same head performs the same role across
models, only that the same \emph{protocol} (discovery/held-out
selection against matched controls) again separates a real,
held-out-replicated effect from the control distributions. We report
this as a second data point for the paper's methodological claim, not
as evidence about a shared or universal ``induction circuit'' across
GPT-2 and DistilGPT2.

\section{Reproducibility details}
\label{app:repro}

All results are produced by \texttt{scripts/run\_neurips\_experiments.py}
and saved to \texttt{neurips/experiment\_results/gpt2\_results.json}
(main results, Table~\ref{tab:main}) and \texttt{distilgpt2\_results.json}
(replication, Table~\ref{tab:distilgpt2}). These files record the device,
torch (2.8.0) and transformers (4.57.3) versions, all random seeds,
per-split example counts, the number of bootstrap
resamples (2{,}000) and control draws (1{,}000), the full per-head
$\Delta$ vector for every split and method, and per-example
log-probabilities for every baseline. Exact model revisions and hashes
are recorded in
\texttt{neurips/experiment\_results/reproducibility\_provenance.md}. A separate
one-time, higher-power
rerun of the main GPT-2 pipeline with $N{=}500$ (rather than 200) per
task, run once and reported regardless of outcome, is saved to
\texttt{gpt2\_n500\_results.json}. Its preserved held-out effect and
weaker specificity contrast are discussed in
Section~\ref{sec:discussion}. Figures are produced by
\texttt{scripts/make\_figures.py}, which reads only the primary JSON
file and performs no additional computation. A claim-to-evidence map
(\texttt{neurips/claim\_evidence\_map.md}) lists, for every quantitative
statement in this paper, the exact JSON field it was read from. Every file
named in this appendix, together with the pipeline source, the figure script,
and a README giving the pinned environment and the exact command for each
reported run, is available at
\url{https://github.com/TheClassicTechno/InterpretabilityNLPProject}. No dataset
is shipped: both task families are generated in memory by
\texttt{src/dataset.py} from the seeds recorded in each result file.

\end{document}